\documentclass[conference,a4paper]{IEEEtran}
\IEEEoverridecommandlockouts
\usepackage{cite}
\usepackage{amsmath,amssymb,amsfonts}
\usepackage{algorithmic}
\usepackage{graphicx}
\usepackage{textcomp}
\usepackage{xcolor}
\usepackage{booktabs}
\usepackage{multirow}
\usepackage{subfig}
\usepackage{graphicx}

\def\BibTeX{{\rm B\kern-.05em{\sc i\kern-.025em b}\kern-.08em
    T\kern-.1667em\lower.7ex\hbox{E}\kern-.125emX}}
\begin{document}

\title{Tuna Nutriment Tracking using Trajectory Mapping in Application to Aquaculture Fish Tank}

\author{\IEEEauthorblockN{Hilmil Pradana}
\IEEEauthorblockA{\textit{Graduate School of Life Science and Systems Engineering} \\
\textit{Kyushu Institute of Technology}\\
Fukuoka, Japan \\
Email: pradana.muhamad-hilmil505@mail.kyutech.jp}
\and
\IEEEauthorblockN{Keiichi Horio}
\IEEEauthorblockA{\textit{Graduate School of Life Science and Systems Engineering} \\
\textit{Kyushu Institute of Technology}\\
Fukuoka, Japan \\
Email: horio@brain.kyutech.ac.jp}
}

\maketitle

\begin{abstract}
The cost of fish feeding is usually around 40 percent of total production cost.
Estimating a state of fishes in a tank and adjusting an amount of nutriments play an important role to manage cost of fish feeding system.
Our approach is based on tracking nutriments on videos collected from an active aquaculture fish farm.
Tracking approach is applied to acknowledge movement of nutriment to understand more about the fish behavior.
Recently, there has been increasing number of researchers focused on developing tracking algorithms to generate more accurate and faster determination of object.
Unfortunately, recent studies have shown that efficient and robust tracking of multiple objects with complex relations remain unsolved.
Hence, focusing to develop tracking algorithm in aquaculture is more challenging because tracked object has a lot of aquatic variant creatures.
By following aforementioned problem, we develop tuna nutriment tracking based on the classical minimum cost problem which consistently performs well in real environment datasets.
In evaluation, the proposed method achieved 21.32 pixels and 3.08 pixels for average error distance and standard deviation, respectively.
Quantitative evaluation based on the data generated by human annotators shows that the proposed method is valuable for aquaculture fish farm and can be widely applied to real environment datasets.

\end{abstract}


\section{Introduction}
Aquaculture is the one of farming type in which aquatic creatures require acceptable environment for living habitat and availability nutriment to increase productivity and sustain healthy growth \cite{Fazio, Freitas, Carmen, Liu}.
Within current requiring acceptable habitat, water quality is also a vital component to enlarge fish fertility rate \cite{Carmen, Liu, Farheen, HLiu}.
Water quality can be obtained by cleaned often and give optimal amount of nutriment.
Increasing number of nutriment can affect a lot of foods wasted in the water and quality of water occurs highly polluted.
On the other hand, reducing feeding will lead starvation and drop fish quality.
So that, management of nutriment delivered is vital component to balance productivity rate \cite{Higuchi, Barron}.

The cost of fish feeding is usually around 40 percent of total production cost \cite{AtoumY, Sabari, Oostlander}.
Estimating a state of fishes in a tank and adjusting an amount of nutriments play an important role to manage cost of fish feeding system.
It is applied to control the amount of nutriment and realizes the fish behavior in tank.
Lately, application to monitor fish behavior has been adopted by a telemetry-based approach \cite{Bridger, Conti} and a computer vision(CV)-based approach \cite{Costa, Xu, Stien, Zion, Duarte, Atoum, Adegboye}.

A telemetry-based approach is a technique attaching an external transmitter by mounting, or surgical implantation in the peritoneal cavity \cite{Bridger}.
Attaching a transmitter in each fish will spend higher cost and its transmitter can only set in large fish.
When their fishes had been farmed, attachment will always be given to new fishes.
On the other hand, CV-based approach studies are not required complexity analysis such as ripple activity and tracking analysis in which, small number of fishes and small tanks with special environment assist creating result.
Tracking approach is applied to acknowledge movement of nutriment to understand more about the fish behavior.
Fish behaviors can be obtained by combination between tracking analysis and ripple activity.
Then, these fish behaviors can be a decision to start and stop fish feeding machine by understanding of ripple activity after giving several nutriments.
By explaining of fish behavior, tracking nutriment is important and it is required to analyze the complexity data in real environment.

Recently, there has been increasing number of researchers focused on developing tracking algorithms to generate more accurate and faster determination of object.
Tracking can be represented as a graph problem which can solved by a frame-by-frame \cite{Ess, Breitenstein, Pellegrini, Ueno} or track-by-track \cite{Berclaz, Zhang2019}.
Interpretation of tracking problems with data association mostly uses a graph, where each detection is called as vertex, and each edge is pointing any possible link among them out as object tracked.
Data association can be declared as minimum cost problem \cite{Berclaz2011, Jiang, Zhang, Pirsiavash} with learning cost problem \cite{Leal} or motion pattern maps \cite{Leal2011}.
Alternative formulations to solve optimization problems is minimum clique problem \cite{Zamir} and lifted multicut problem \cite{Tang} where its formulations follow body pose layout to obtain estimated model.
Recently, efficient and robust tracking of multiple objects with complex relations remain unsolved.
Hence, focusing to develop tracking algorithm in aquaculture is more challenging because tracked object has a lot of aquatic variant creatures.
By summarizing aforementioned problems, we proposed tuna nutriment tracking based on the classical minimum cost problem \cite{Jiang, Zhang, Pirsiavash} where each detection calculates minimum distance among them and creates a trajectory to be tracked line.
By collaborating with an active aquaculture fish farm, we develop tuna nutriment tracking using trajectory mapping.
A video camera is placed above the boat with a highly disturbance of ocean wave and many dense nutriments.
The camera captures between ocean surface and fish feeding machine.
After that, videos transfer to a computer for further analysis the behavior of fish.

The aim of this research is tracking approach to acknowledge the behavior of tuna.
For next, it can be useful to improve the production profit in fish farms by controlling the amount of nutriment in optimal rate.

\noindent To summarize, we make the following \textbf{contributions}:
\begin{itemize}
\item We propose tuna nutriment tracking based on trajectory mapping which can perform well as well as human annotator results.
\item We propose a new novel small nutriment tracking method with collecting information of leading line into ripple.
\item We show significantly improvement result of trajectory mapping in real environment datasets.
\end{itemize}

\section{The Proposed Method - Trajectory Mapping}
\label{section:Methods}

\begin{figure*}[t!]
  \centering
    \includegraphics[width=0.89\textwidth]{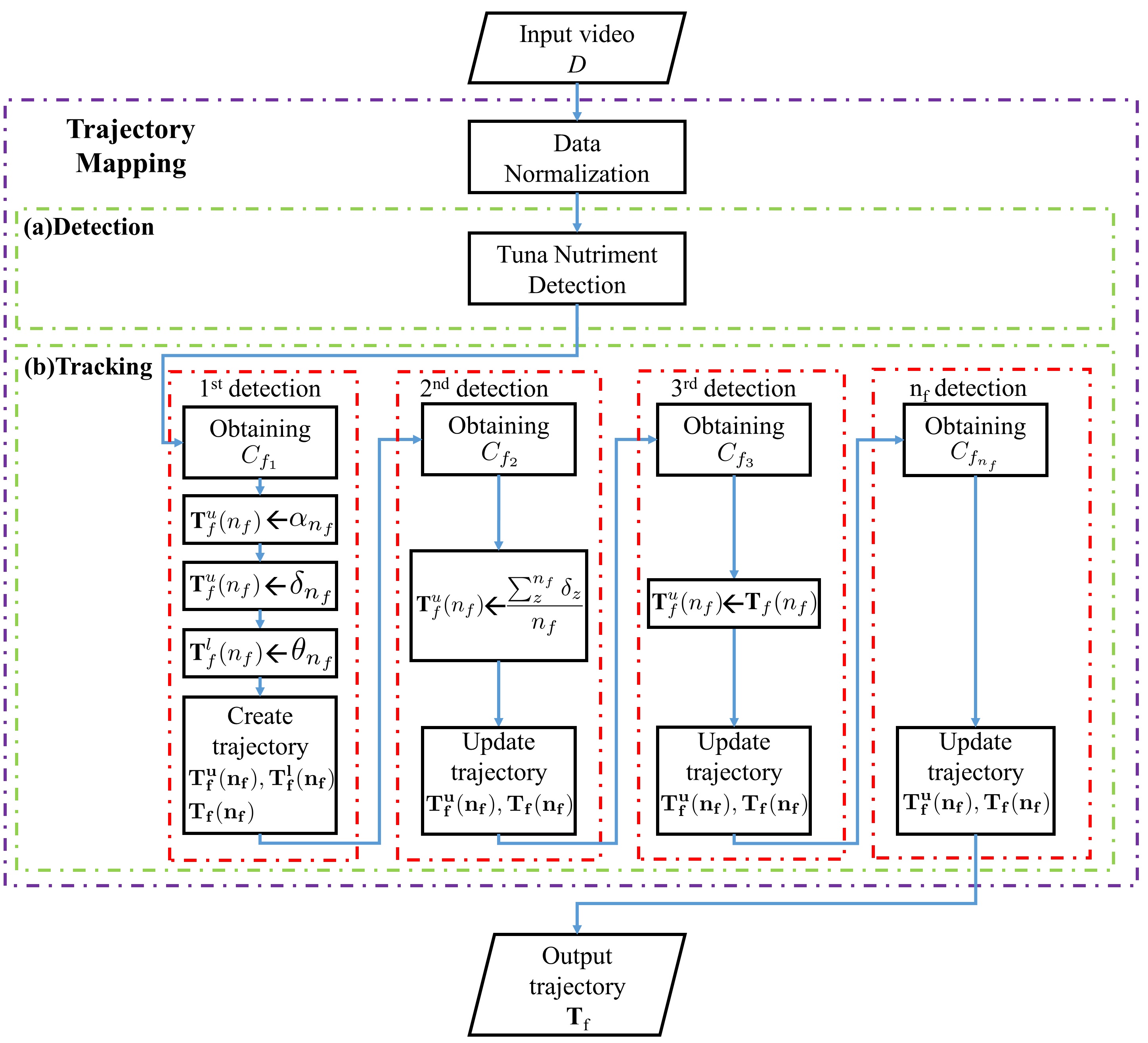}
    \caption{
    Flowchart of the proposed trajectory mapping.
    The input video is received and applied image stabilization as data normalization.
    (a) Creating model for tuna nutriment detection using YOLOv3 \cite{Redmon} and obtaining bounding box for each tuna nutriment prediction in all frames of a video.
    (b) Tuna nutriment prediction as input tracking approach to be initialization for $n_f = 1$ detection.
    After that, we obtain upper and lower limit trajectory $\mbox{\boldmath$ T$}^u_l$, $\mbox{\boldmath$ T$}^l_f$ and the maximum height of all tuna nutriment predictions $\delta_1$.
    Next, we find the value of $C_{f_2}$ using the shortest distance between $C_{f_1}$ and all tuna nutriment predictions in $n_f$ appearing in inside area between $\mbox{\boldmath$ T$}^u_f(1)$ and $\mbox{\boldmath$ T$}^u_f(1)$.
    Its process is repeatedly until $n_f$.
    Next, we obtain final trajectory as tracking result.
    }
    \label{fig:figure21}
\end{figure*}

\begin{figure}[t!]
  \centering
    \includegraphics[width=0.49\textwidth]{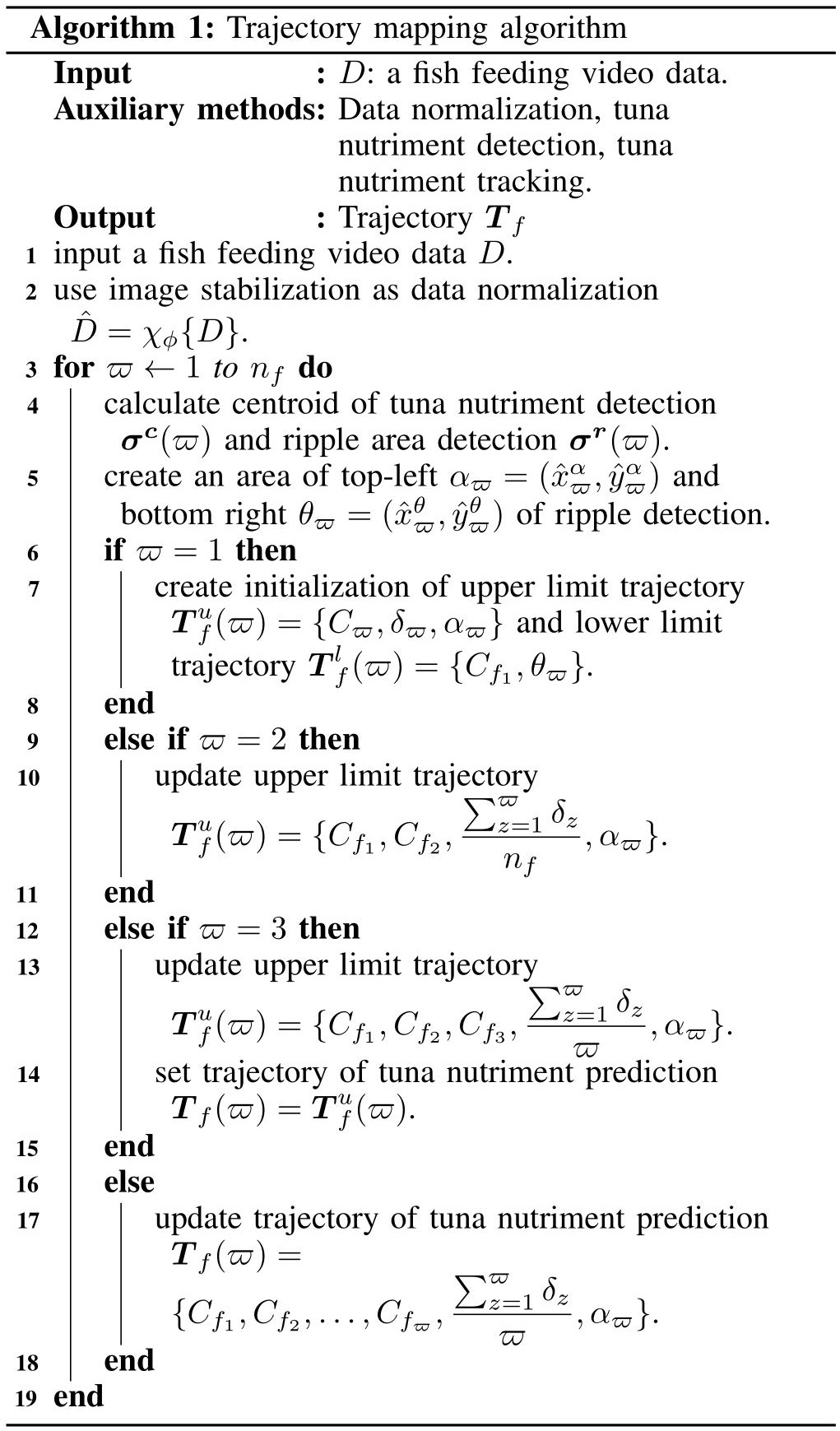}
    \caption{
    Algorithm of the proposed trajectory mapping.
    Iterative method are applied to improve trajectory $\mbox{\boldmath$ T$}_f$ by collecting each centroid of nutriment in every frame.
    }
    \label{fig:figure22}
\end{figure}

Our formulation is based on the classical minimum cost problem where each detection calculates minimum distance among them and creates a trajectory to be tracked line.
In order to provide some background and formally introduce our approach, we start by providing flowchart and algorithm of tuna nutriment tracking.
We then explain how the proposed method works to real environment.
The proposed trajectory mapping contains a data normalization process, tuna nutriment detection and tuna nutriment tracking.
The system flowchart of the proposed method is shown in Fig. \ref{fig:figure21}, and the algorithm of the proposed trajectory mapping is represented in Fig. \ref{fig:figure22} where $D$ and $\textbf{T}_f$ are input video and trajectory of time-ordered tuna nutriment, respectively.
\subsection{Data Normalization}
\label{section:DataNormalization}
For data normalization, image stabilization is applied to reduce a hand-held camera and ocean waves.
Image stabilization is created by transformation from previous to current frame using optical flow for all frames.
\cite{Nghia} accumulates rigid transformation $\chi$ to obtain linked between frame $L$.
New rigid transformation $\chi_\phi$ in frame $\phi$ can be written as:

\begin{equation}\label{E:Equation21}
\begin{split}
\chi_\phi =& \chi_{\phi-1} + (\dfrac{1}{\gamma} \sum_{\tau=\phi-\gamma}^{\phi+\gamma} L_\tau) – L_{\phi-1}, \\
\hat{D} =& \chi_\phi\{D\},
\end{split}
\end{equation}

\noindent where $\hat{D}$ is output video after applied image stabilization and $\gamma$ is smoothing radius where the radius is number of frames used for smoothing and defined by 30.

\subsection{Tuna Nutriment Detection}
\label{section:SmallFishesDetection}
The idea of tuna nutriment detection is to produce boundary box in each nutriment associated in tracking method.
In implementation of tuna nutriment detection, YOLOv3 \cite{Redmon} accumulates bounding box of tuna nutriment prediction $B = (\hat{x}, \hat{y}, \hat{w}, \hat{h})$ by training model with bounding box $P = (p_x$, $p_y$, $p_w$, $p_h)$ of ground truth data where $p_x$, $p_y$, $p_w$, and $p_h$ are centroid $x$, centroid $y$, width, and height of bounding box in ground truth data, respectively.
$\varsigma_x$ and $\varsigma_y$ represent the absolute location of the top-left corner of the current grid cell. $w$ and $h$ are the absolute width and height to the whole image.
Bounding box of tuna nutriment prediction $B$ can defined as:

\begin{equation}\label{E:Equation22}
\begin{split}
\hat{x} &= \delta(p_x) + \varsigma_x \\
\hat{y} &= \delta(p_y) + \varsigma_y \\
\hat{w} &= e^{p_w} * w \\
\hat{h} &= e^{p_h} * h
\end{split}
\end{equation}

where $\delta$ is model followed by \cite{Redmon}.

\subsection{Tuna Nutriment Tracking}
\label{section:SmallFishesTracking}
In order to represent tracking of tuna nutriment, introducing how to collect set of tuna nutriment prediction corresponding to time-ordered path in the graph is important.
We are given $\mbox{\boldmath$ \sigma^c$}(n) = \{C_1, C_2, C_3, \ldots, C_n\}$ as input centroid of tuna nutriment predictions where $n$ is the total number of nutriment for all frames of video $\hat{D}$.
Each tuna nutriment prediction is represented by $C_n = \{\hat{x}^c_n, \hat{y}^c_n\}$.
Definition of a trajectory is denoted as centroid of time-ordered tuna nutriment predictions $\mbox{\boldmath$ T$}_f(n_f) = \{C_{f_1}, C_{f_2}, C_{f_3}, \ldots, C_{f_{n_f}}\}$ where $n_f$ is the number of detections formed by trajectory $f$.
So that, $\varrho = \{n_1, n_2, n_3, \ldots, n_{n_f}\}$ can be denoted as the total of number of nutriments appearing in every time-ordered trajectory $\mbox{\boldmath$ T$}_f(n_f)$.

\subsubsection{Problem Statement}
\begin{figure}[t!]
  \centering
    \includegraphics[width=0.44\textwidth]{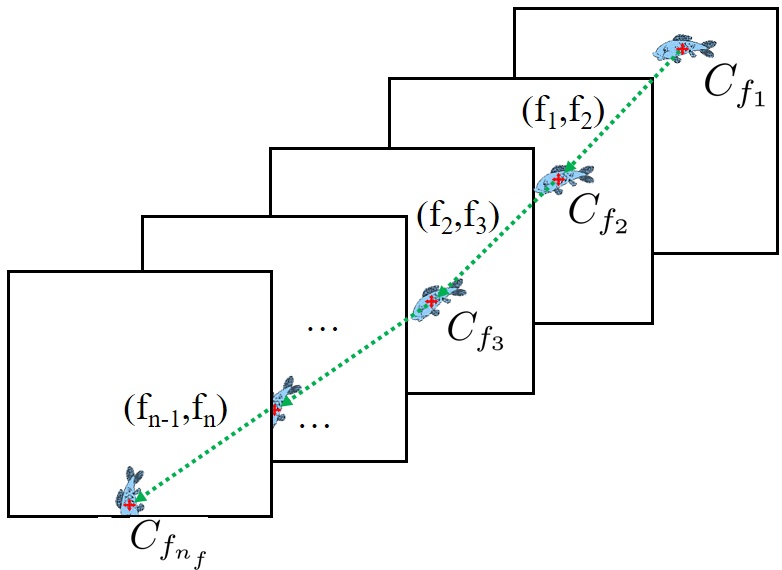}
    \caption{
    Visualization of trajectory of tuna nutriment predictions $\mbox{\boldmath$ T$}_f = \{C_{f_1}, C_{f_2}, C_{f_3}, \ldots, C_{f_{n_f}}\}$ in which every tuna nutriment predictions is connected by node $\{(f_1, f_2),(f_2, f_3), \ldots, (f_{n-1_f}, f_{n_f})\}$.
    }
    \label{fig:figure23}
\end{figure}

The problem can be represented with an undirected graph $G = (V,E)$, where $V := \{1,...,n\}, E \subset V^{2}$, and each node $f \in V$ denotes a unique detection $C_f \in \sigma^c$.
The task of dividing the set of tuna nutriment predictions into trajectories can be observed as grouping nodes in graph.
Fig. \ref{fig:figure23} shows that each trajectory $\mbox{\boldmath$ T$}_f(n_f) = \{C_{f_1}, C_{f_2}, C_{f_3}, \ldots, C_{f_{n_f}}\}$ in the scene can be mapped into a group of nodes $\{(f_1, f_2),(f_2, f_3), \ldots, (f_{n-1_f}, f_{n_f})\}$.
To produce each $C_{f_{[1, n_f]}}$, trajectory mapping is applied in next section.

In two-dimensional trajectory, the component of trajectory is divided by horizontal and vertical direction.
In vertical direction, acceleration is constant and has quadratic function.
Trajectory mapping applies the idea of acceleration and chooses quadratic function as basis.

To produce quadratic function $y^c = a^c_3x^2 + a^c_2x + a^c_1$ as a result of trajectory $\mbox{\boldmath$ T$}_f$, we apply polynomial fitting \cite{Weisstein} defined by calculation of $\hat{x}^c_{n_f}$ to form Vandermonde matrix $\mbox{\boldmath$ V$}$ with $3$ columns as results of $\mbox{\boldmath$ a$}^c$.

\begin{equation}\label{E:Equation221}
\begin{split}
\begin{bmatrix}
     1	& \hat{x}^c_1		& (\hat{x}^c_1)^2 \\
     1	& \hat{x}^c_2 		& (\hat{x}^c_2)^2 \\
     1	& \vdots			& \vdots \\
     1	& \hat{x}^c_{n_f}	& (\hat{x}^c_{n_f})^2
\end{bmatrix}
\begin{bmatrix}
    a^c_1 \\
    a^c_2 \\
    a^c_3
\end{bmatrix}
=
\begin{bmatrix}
    \hat{y}^c_1\\
    \hat{y}^c_2\\
    \vdots\\
    \hat{y}^c_{n_f}
\end{bmatrix}
\end{split}
\end{equation}

\noindent (\ref{E:Equation221}) can be inverted directly.
To yield the solution vector $\mbox{\boldmath$ a$}^c$, it can be defined as: 

\begin{equation}\label{E:Equation222}
\begin{split}
\mbox{\boldmath$ a$}^c = (\mbox{\boldmath$ X$}^T\mbox{\boldmath$ X$})^{-1}\mbox{\boldmath$ X$}^T\mbox{\boldmath$ Y$},
\end{split}
\end{equation}

\subsubsection{Tuna Nutriment Predictions $C_{f_{n_f}}$ where $n_f = 1$ as Initialization Point Detection}
\label{section:nf1}
Tuna nutriment predictions $C_{f_1}$ are obtained from every tuna nutriment prediction $C_n$ in around cutting area of $w$.
To define cutting area, we use centroid $\hat{x}_{f_1}$ as component of $C_{f_1}$ by thresholding in $w$ which is defined as:

\begin{equation}\label{E:Equation23}
\begin{split}
w*\gamma \leq \hat{x}_{f_1} \leq w,
\end{split}
\end{equation}

\noindent where $\gamma$ is an input parameter and empirically defined as $0.9$.

Direction of nutriment is calculated by leading nutriment to ripple area around sea levels.
We are given a pair set of ripple area detection $\mbox{\boldmath$ \sigma^r$}(n_f) = \{(R_{11}, R_{12}), (R_{21}, R_{22}), (R_{31}, R_{32}), \ldots, (R_{n_f1}, R_{n_f2})\}$ as time-ordered ripple predictions in number of detections $n_f$.
Each ripple prediction is represented by $R_{n_f1,2} = \{\hat{x}^r_{n_f}, \hat{y}^r_{n_f}, \hat{w}^r_{n_f}, \hat{h}^r_{n_f}\}$.
We divide component of ripple prediction to be an area of top-left $\alpha_{n_f} = (\hat{x}^{\alpha}_{n_f}, \hat{y}^{\alpha}_{n_f})$ and bottom right $\theta_{n_f} = (\hat{x}^{\theta}_{n_f}, \hat{y}^{\theta}_{n_f})$ of ripple detection by following:

\begin{equation}\label{E:Equation24}
\begin{split}
\hat{x}^\alpha_{n_f} =& \hat{x}^r_{n_f} - \dfrac{\hat{w}^r_{n_f}}{2}, \\
\hat{y}^\alpha_{n_f} =& \hat{y}^r_{n_f} - \dfrac{\hat{h}^r_{n_f}}{2}, \\
\hat{x}^\theta_{n_f} =& \hat{x}^r_{n_f} + \dfrac{\hat{w}^r_{n_f}}{2}, \\
\hat{y}^\theta_{n_f} =& \hat{y}^r_{n_f} + \dfrac{\hat{h}^r_{n_f}}{2},
\end{split}
\end{equation}

To obtain more feature, we need to know possibly coverage area for possibly nutriment appearing in next frame by creating upper and lower limit trajectory $\mbox{\boldmath$ T$}^u_f$ and $\mbox{\boldmath$ T$}^l_f$, respectively.
Upper and lower limit trajectory $\mbox{\boldmath$ T$}^u_f(n_f)$ and $\mbox{\boldmath$ T$}^l_f(n_f)$ formed by trajectory $f$ are initialized by following:

\begin{equation}\label{E:Equation25}
\begin{split}
\mbox{\boldmath$ T$}^u_f(n_f) =& \{C_{f_1}, \delta_{n_f}, \alpha_{n_f}\}, \\
\mbox{\boldmath$ T$}^l_f(n_f) =& \{C_{f_1}, \theta_{n_f}\}, \\
\end{split}
\end{equation}

\noindent where $\delta$ is the maximum height of all nutriment detections in $n_f$.
(\ref{E:Equation25}) can be simplify by substituting $n_f = 1$ to be:

\begin{equation}\label{E:Equation26}
\begin{split}
\mbox{\boldmath$ T$}^u_f(1) =& \{C_{f_1}, \delta_1, \alpha_{1}\}, \\
\mbox{\boldmath$ T$}^l_f(1) =& \{C_{f_1}, \theta_{1}\}, \\
\end{split}
\end{equation}

\noindent where $\delta_1 = (\dfrac{\hat{x}^\alpha_1 + w}{2}, \min\limits_{1 \leq z \leq n_1} \hat{y}^c_z)$.

\subsubsection{Tuna Nutriment Predictions $C_{f_{n_f}}$ where $n_f = 2$}
\label{section:nf2}
To be a candidate of $C_{f_2}$, we use all tuna nutriment predictions $n$ appearing in the inside of area between $y^u = a^u_3x^2 + a^u_2x + a^u_1$ and $y^l = a^l_3x^2 + a^l_2x + a^l_1$.
Vector $\mbox{\boldmath$ a$}^u$ and $\mbox{\boldmath$ a$}^l$ are produced by calculating $\mbox{\boldmath$ T$}^u_f({n_f} - 1)$ and $\mbox{\boldmath$ T$}^l_f({n_f} - 1)$ with Vandermonde matrix shown in (\ref{E:Equation221}) and (\ref{E:Equation222}), respectively.
Given $\mbox{\boldmath$ \sigma^\kappa$}(\mu) = \{\kappa_{1_{n_f}}, \kappa_{2_{n_f}}, \kappa_{3_{n_f}}, \ldots, \kappa_{\mu_{n_f}}\}$ is a set of candidate $C_{f_{n_f}}$.
$C_{f_{n_f}}$ is defined by the nutriment predictions which have shortest distance denoted by:
%


\begin{eqnarray}\label{E:Equation27}
C_{f_{n_f}}&{}={}&\operatorname*{arg\,min}_{\mu} (\mbox{\boldmath$ Z$}(f_{n_f-1}) - \mbox{\boldmath$ \sigma^\kappa$}(\mu))^T\nonumber\\
&&(\mbox{\boldmath$ Z$}(f_{n_f-1}) - \mbox{\boldmath$ \sigma^\kappa$}(\mu))
\end{eqnarray}


\noindent where $\mbox{\boldmath$ Z$}(f_{n_f-1}) = \{C^1_{f_{n_f-1}}, C^2_{f_{n_f-1}}, C^3_{f_{n_f-1}}, \ldots, C^\mu_{f_{n_f-1}}\}$.
(\ref{E:Equation27}) can be simplify to be;

\begin{equation}\label{E:Equation28}
\begin{split}
C_{f_2} = \operatorname*{arg\,min}_{\mu} (\mbox{\boldmath$ Z$}(1) - \mbox{\boldmath$ \sigma^\kappa$}(\mu))^T(\mbox{\boldmath$ Z$}(1) - \mbox{\boldmath$ \sigma^\kappa$}(\mu)),
\end{split}
\end{equation}


\noindent Updating upper trajectory $\mbox{\boldmath$ T$}^u_f(n_f)$ can be defined as:

\begin{equation}\label{E:Equation210}
\begin{split}
\mbox{\boldmath$ T$}^u_f(n_f) =& \{C_{f_1}, C_{f_2}, \dfrac{\sum_{z=1}^{n_f} \delta_z}{n_f}, \alpha_{n_f}\},
\end{split}
\end{equation}

\subsubsection{Tuna Nutriment Predictions $C_{f_{n_f}}$ where $n_f = 3$}
\label{section:nf3}
Minimum requirement for trajectory of quadratic functions must have at least 3 tuna nutriment predictions collected.
To produce $C_{f_3}$, (\ref{E:Equation27}) is applied using $n_f = 3$ as parameter.
Then, updating upper limit trajectory $\mbox{\boldmath$ T$}^u_f(n_f)$ is denoted as follows:

\begin{equation}\label{E:Equation211}
\begin{split}
\mbox{\boldmath$ T$}_f(n_f) =& \mbox{\boldmath$ T$}^u_f(n_f), \\
\end{split}
\end{equation}

\subsubsection{Tuna Nutriment Predictions $C_{f_{n_f}}$ where $n_f \geq 4$}
\label{section:nf4}
To precise accuracy of trajectory $\mbox{\boldmath$ T$}_f$, we refine its trajectory by collecting more tuna nutriment prediction $C_{f_{n_f}}$.
Tuna nutriment prediction $C_{f_{n_f}}$ is calculated using the nearest nutriment detection in area of $y^u = a^u_3x^2 + a^u_2x + a^u_1$ with tolerance degree from quadratic function between $\pm 30$ degree.

To handle losing tuna nutriment prediction, we used previously tuna nutriment prediction by calculating the speed of nutriment in next frame.

\begin{equation}\label{E:Equation212}
\begin{split}
\hat{x}^c_{f_{n_f}} =& 3\hat{x}^c_{f_{n_f - 1}} - 3\hat{x}^c_{f_{n_f - 2}} + \hat{x}^c_{f_{n_f - 3}}, \\
\hat{y}^c_{f_{n_f}} =& a^c_3(\hat{x}^c_{f_{n_f}})^2 + a^c_2\hat{x}^c_{f_{n_f}} + a^c_1,
\end{split}
\end{equation}

\noindent where $a^c_1$, $a^c_2$, and $a^c_3$ are coefficients of quadratic function formed by trajectory $\mbox{\boldmath$ T$}_f$

\section{Experiment}
In this section, we first explain the details of our datasets.
We then describe evaluation approach to calculate error rate distance and show quantitative evaluation with various of $n_f$ to discover an optimal value.

\subsection{Datasets}
We report our datasets containing 1 video which has interference of hand-held camera and ocean waves with 419 frames.
Each dimension of frame has $1920\times1080$ pixels. 
Its video sequences are in MOV format with frame rate 30 frames/second.
Range size of nutriment is starting from $9 \times 6$ to $13 \times 36$ pixels.

\subsection{Evaluation Approach}
Evaluation approach is defined by measuring minimum euclidean distance based on number of nutriment collected with ground truth $\mbox{\boldmath$ T$}_g$.
Best trajectory $T^*$ with minimum error rate distance is defined as:

\begin{equation}\label{E:Equation32}
\begin{split}
T^* =  \operatorname*{arg\,min}_{n_f}(\mbox{\boldmath$ T$}_g - \mbox{\boldmath$ T$}(n_f))^T(\mbox{\boldmath$ T$}_g - \mbox{\boldmath$ T$}(n_f)),
\end{split}
\end{equation}

\noindent where $n_f \in [3,9]$.

\subsection{Quantitative Evaluation with various of $n_f = [3,9]$}

\begin{figure}[t!]
  \centering
    \includegraphics[width=0.49\textwidth]{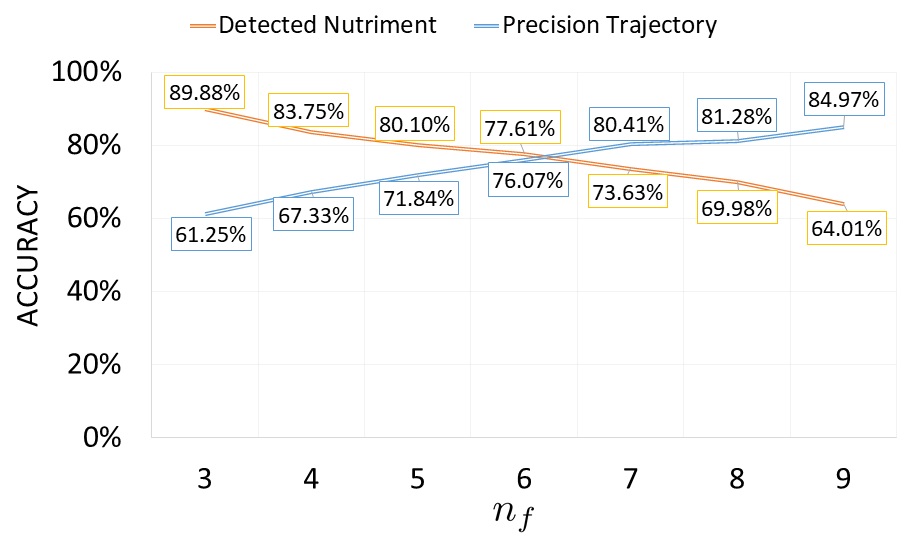}
    \caption{
	Accuracy detected nutriment and precision trajectory in various $n_f \in [3,9]$.
	The optimum value of detected nutriment and precision trajectory is $n_f = 6$.
    }
    \label{fig:figure31}
\end{figure}

\begin{figure}[t!]
  \centering
    \includegraphics[width=0.45\textwidth]{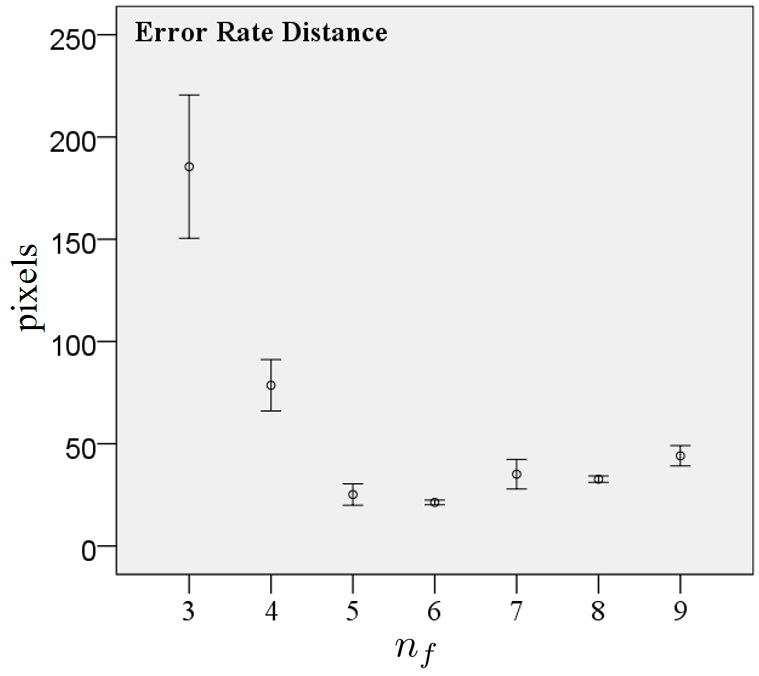}
    \caption{
	95\% Confidence interval of error rate distance in various of $n_f \in [3,9]$.
	The lowest error rate distance of proposed method is $n_f = 6$.
    }
    \label{fig:figure32}
\end{figure}

\begin{table}[]
\centering
\caption{Statistical analysis of various $n_f \in [3,9]$ for quantitative evaluation using one-samples T-Test.}
\label{Table:Table1}
\begin{tabular}{@{}lllllll@{}}
\toprule
\multicolumn{1}{c}{\multirow{2}{*}{\textbf{nf}}} & \multicolumn{1}{c}{\multirow{2}{*}{n}} & \multicolumn{1}{c}{\multirow{2}{*}{\begin{tabular}[c]{@{}c@{}}Mean\\ (pixels)\end{tabular}}} & \multicolumn{1}{c}{\multirow{2}{*}{\begin{tabular}[c]{@{}c@{}}Std. Dev.\\ (pixels)\end{tabular}}} & \multicolumn{1}{c}{\multirow{2}{*}{\begin{tabular}[c]{@{}c@{}}Std. Error\\ (pixels)\end{tabular}}} & \multicolumn{2}{c}{\begin{tabular}[c]{@{}c@{}}95\%   Confidence\\ Interval of the Diff.\end{tabular}}                                                      \\ \cmidrule(l){6-7} 
\multicolumn{1}{c}{}                             & \multicolumn{1}{c}{}                   & \multicolumn{1}{c}{}                                                                         & \multicolumn{1}{c}{}                                                                              & \multicolumn{1}{c}{}                                                                               & \multicolumn{1}{c}{\begin{tabular}[c]{@{}c@{}}Lower\\ (pixels)\end{tabular}} & \multicolumn{1}{c}{\begin{tabular}[c]{@{}c@{}}Upper\\ (pixel)\end{tabular}} \\ \midrule
3                                                & 30                                     & 185.47                                                                                       & 93.81                                                                                             & 17.13                                                                                              & 150.44                                                                       & 220.5                                                                       \\
4                                                & 30                                     & 78.58                                                                                        & 33.64                                                                                             & 6.14                                                                                               & 66.02                                                                        & 91.14                                                                       \\
5                                                & 30                                     & 25.17                                                                                        & 13.97                                                                                             & 2.55                                                                                               & \textbf{19.96}                                                               & 30.39                                                                       \\
6                                                & 30                                     & \textbf{21.32}                                                                               & \textbf{3.08}                                                                                     & \textbf{0.56}                                                                                      & 20.18                                                                        & \textbf{22.48}                                                              \\
7                                                & 30                                     & 35.12                                                                                        & 19.34                                                                                             & 3.53                                                                                               & 27.9                                                                         & 42.35                                                                       \\
8                                                & 30                                     & 32.67                                                                                        & 4.19                                                                                              & 0.76                                                                                               & 31.11                                                                        & 34.24                                                                       \\
9                                                & 30                                     & 44.1                                                                                         & 13.32                                                                                             & 2.43                                                                                               & 39.12                                                                        & 49.07                                                                       \\ \bottomrule
\end{tabular}
\end{table}

Quantitative evaluation is computed by performance of detected nutriment and precision trajectory showed in Fig. \ref{fig:figure31}.
Number of detected nutriment is defined as percentage of detected nutriment divided by ground truth of nutriments appearing in frame.
Meanwhile, precision trajectory is computed by total number of nutriments having trajectory leading to ripple area divided by detected nutriment.
Fig. \ref{fig:figure32} and Table \ref{Table:Table1} show the confidence interval and statistical analysis of error rate distance in various of $n_f$.
The results show that the optimal value of $n_f$ is $6$ in which this parameter produces smallest error rate distance.

\section{Result}
In this section, we compare proposed method and state-of-the-art benchmark methods on our datasets.
After that, we show the figures to explain the advantage of the proposed method and computational time between proposed method and state-of-the-art benchmark methods.

\subsection{Evaluation Result}

\begin{figure}[t!]
  \centering
    \includegraphics[width=0.49\textwidth]{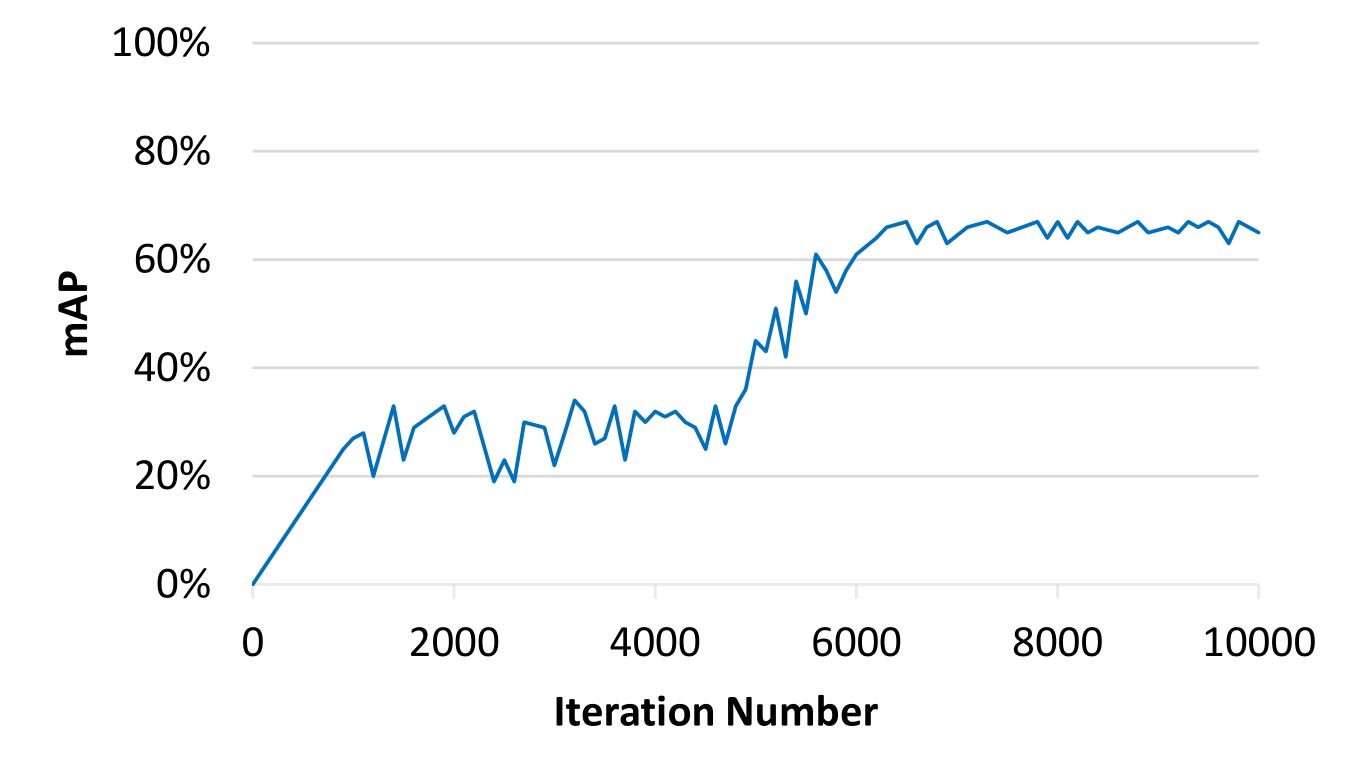}
    \caption{
    mAP result during training model using YOLOv3 \cite{Redmon}.
    After 7k iterations, curve of mAP of model has more stable. 
    }
    \label{fig:figure41}
\end{figure}

\begin{figure*}[!t]
\centerline{\subfloat[Trajectory Mapping]{\includegraphics[width=0.37\textwidth]{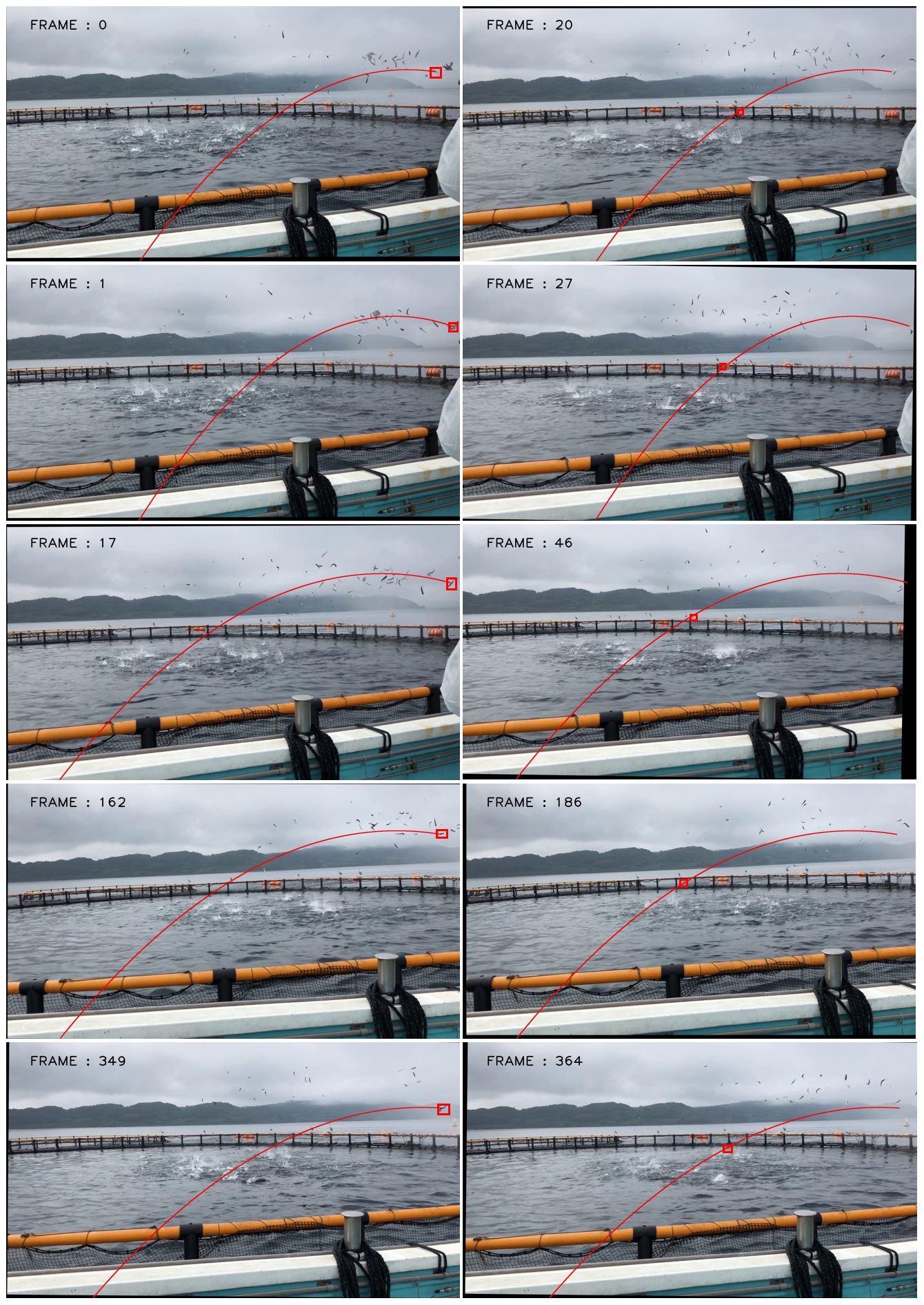}
\label{fig:figure42}}
\hfil
\subfloat[JDE]{\includegraphics[width=0.37\textwidth]{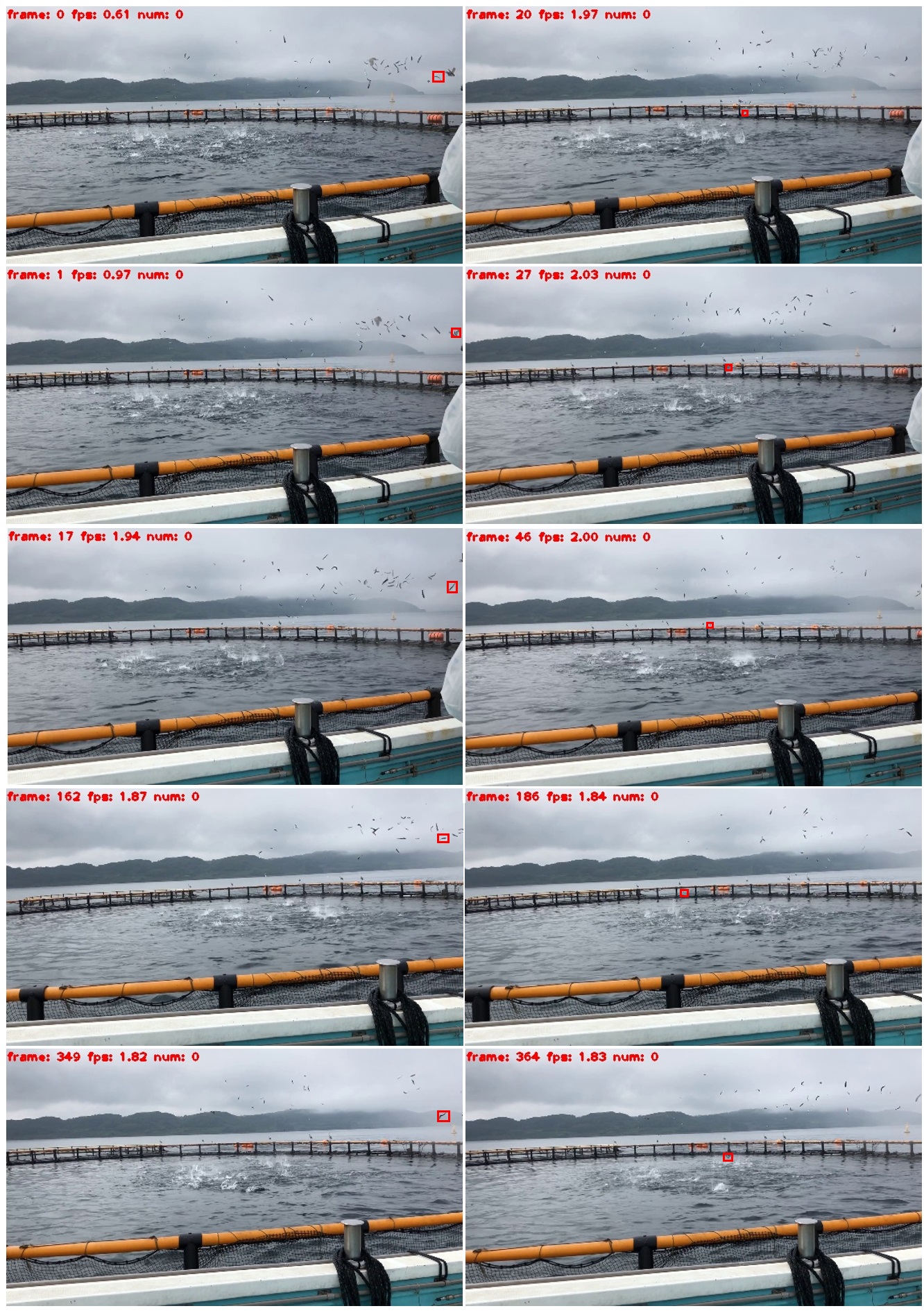}
\label{fig:figure43}}}
\vfil
\centerline{\subfloat[YOLOv3+SORT]{\includegraphics[width=0.37\textwidth]{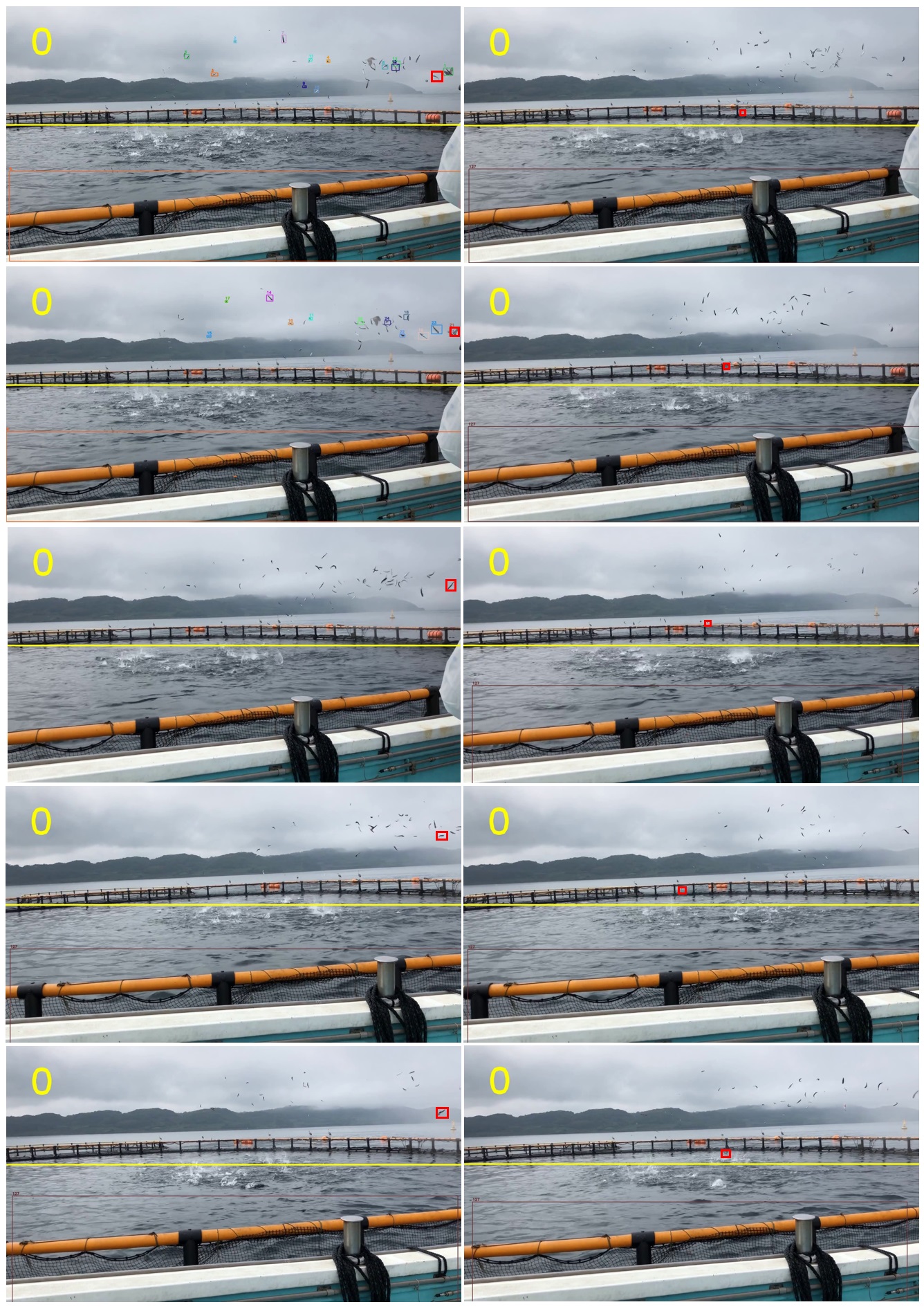}
\label{fig:figure44}}
\hfil
\subfloat[Our Detection Model + SORT]{\includegraphics[width=0.37\textwidth]{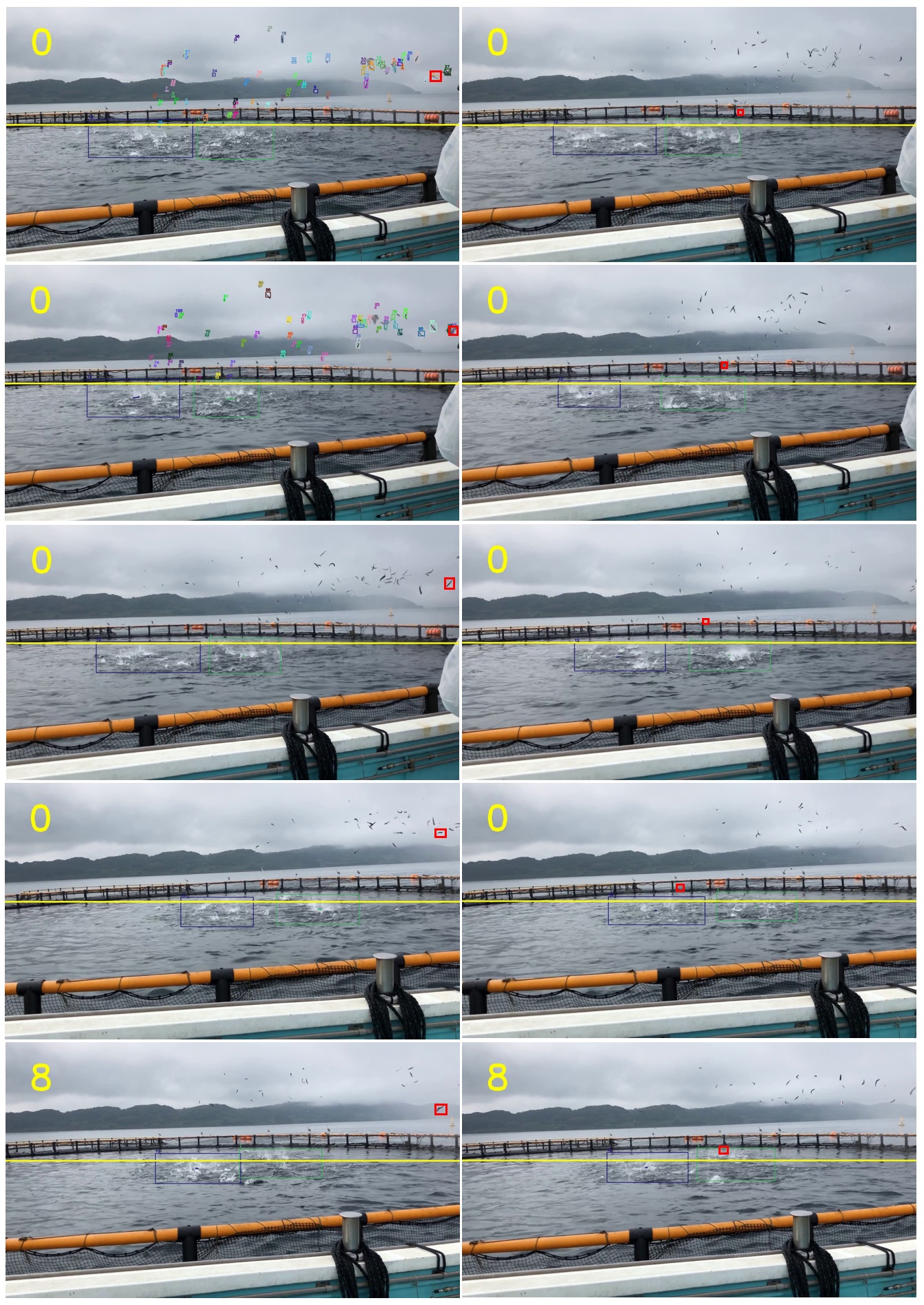}
\label{fig:figure45}}}
\caption{
Observation of proposed method namely trajectory mapping shown in (a) and benchmark method results shown in (b), (c), and (d).
    Left and right images for each method represent first node of nutriment and nutriment tracked after several frames, respectively.
    In trajectory mapping method, red curve is defined as trajectory result of proposed method.
    Red box in each image represents ground truth of nutriment.
    We can see that both red curve and red box in trajectory mapping show precisely tracked result and it proves that trajectory mapping creates trajectory very well while benchmark methods perform poor without tracking results of nutriments even SORT is able to detect some nutriments.}
\label{fig:figure42345}
\end{figure*}

Precision of mAP in object detection is computed by performance YOLOv3 \cite{Redmon} to train our datasets with 10k iterations with $416\times416$ pixels for image resizing from $1920\times1080$ pixels.
Fig. \ref{fig:figure41} displays training result of our datasets using YOLOv3 \cite{Redmon} and reaches 67\% of maximum mAP with 10k iterations.
We also tested proposed methods and state-of-the-art benchmark results.
There are many state-of-the-art methods using multiple object tracking (MOT) \cite{ZWang, WLin, STang, YZhang, FYu, MBabaee, BPang}.
These methods perform well using six publicly available datasets on pedestrian detection, MOT and person search provided by \cite{AMilan, LLealTaixe, AEss}.
In evaluations, we choose JDE \cite{ZWang} to represent MOT as benchmark method because JDE is very fast and accurate based on re-implementation of faster object detection compared with \cite{WLin, STang, YZhang, FYu, MBabaee, BPang}.
We also use SORT \cite{Bewley} as benchmark methods and add our detection model to completely understand performance of tracking method.

In Fig. \ref{fig:figure42}, the proposed method is demonstrated to be able to track small nutriment while JDE and SORT with original YOLOv3 and our detection model perform poor (Fig. \ref{fig:figure43}, \ref{fig:figure44}, \ref{fig:figure45}) without tracking results of nutriments even SORT is able to detect some nutriments.
By our experiment, the benchmark methods fail to run our datasets because the size of nutriment is too small (maximum size is $13 \times 36$ pixels) and the speed of nutriment is fast (average nutriment movement from start to end node is $23.8$ frames).

\subsection{Implementation Details}
\begin{table}[]
\centering
\caption{Hardware and software environment for running proposed method and benchmark methods for comparison.}
\label{Table:Table2}
\begin{tabular}{@{}lll@{}}
\toprule
\multicolumn{3}{c}{Spesification}                                                                                                    \\ \midrule
\multirow{3}{*}{Hardware} & CPU      & Intel Core i7-9700   CPU @3.00GHz (8 CPUs)                                                    \\
                          & RAM      & 16 GB                                                                                         \\
                          & GPU      & NVIDIA GeForce GTX 745                                                                        \\ \midrule
\multirow{3}{*}{Software} & OS       & Windows 10 Pro 64-bit                                                                         \\
                          & IDE      & \begin{tabular}[c]{@{}l@{}}Microsoft Visual Studio Professional 2017\\ v.15.9.25\end{tabular} \\
                          & Language & Python 3.6 64bit                                                                              \\ \bottomrule
\end{tabular}
\end{table}

\begin{table}[]
\centering
\caption{Comparison computational time proposed method and benchmark methods.}
\label{Table:Table3}
\begin{tabular}{lcrrrrr}
\hline
\multicolumn{1}{c}{\multirow{2}{*}{Methods}}                   & \multirow{2}{*}{N}      & \multicolumn{1}{c}{\multirow{2}{*}{\begin{tabular}[c]{@{}c@{}}Mean\\ (fps)\end{tabular}}} & \multicolumn{1}{c}{\multirow{2}{*}{\begin{tabular}[c]{@{}c@{}}Std.\\ Dev.\\ (fps)\end{tabular}}} & \multicolumn{1}{c}{\multirow{2}{*}{\begin{tabular}[c]{@{}c@{}}Std.\\ Err.\\ (fps)\end{tabular}}} & \multicolumn{2}{c}{\begin{tabular}[c]{@{}c@{}}95\% Confidence\\ Interval of the Diff.\end{tabular}}                                                   \\ \cline{6-7} 
\multicolumn{1}{c}{}                                           &                         & \multicolumn{1}{c}{}                                                                      & \multicolumn{1}{c}{}                                                                             & \multicolumn{1}{c}{}                                                                             & \multicolumn{1}{c}{\begin{tabular}[c]{@{}c@{}}Lower\\ (fps)\end{tabular}} & \multicolumn{1}{c}{\begin{tabular}[c]{@{}c@{}}Upper\\ (fps)\end{tabular}} \\ \hline
Ours                                                           & 419                     & 1.93                                                                                      & 0.61                                                                                             & 0.03                                                                                             & 1.87                                                                      & 1.99                                                                      \\
JDE                                                            & \multicolumn{1}{l}{419} & 1.87                                                                                      & 0.07                                                                                             & 0.00                                                                                             & 1.86                                                                      & 1.87                                                                      \\
YOLOv3 + Sort                                                  & \multicolumn{1}{r}{419} & 0.45                                                                                      & -                                                                                                & -                                                                                                & -                                                                         & -                                                                         \\
\begin{tabular}[c]{@{}l@{}}Our Detection +\\ Sort\end{tabular} & \multicolumn{1}{r}{419} & 0.47                                                                                      & -                                                                                                & -                                                                                                & -                                                                         & -                                                                         \\ \hline
\end{tabular}
\end{table}

For analysis of the computational complexity and execution time of the proposed methodology,  a computational time analysis is conducted using a video with 419 frames.
Table \ref{Table:Table2} shows the specification of hardware and software for comparison.
Table \ref{Table:Table3} compares the computation time (in fps) for proposed method, namely trajectory mapping and benchmark approaches: JDE and SORT with original YOLOv3 and our detection model.
For average and standard deviation of computational time, we reach $1.93$  and $0.61$ fps, while JDE spends $1.87$ and $0.07$ fps, respectively.
SORT only provides average computational time without information of computational time for individual frame.
Computational time for both detection model of YOLOv3 and our detection model with SORT performs worst and these benchmark approaches reach $0.45$ fps and $0.47$, respectively.
By analyzing computational complexity, proposed method runs faster than JDE with the different speed is $0.6$ fps.

\section{Conclusion and Discussion}
Tracking approach is the one of features to analyze fish behavior to create a decision to optimize the amount of nutriment.
Recent studies have shown that it is possible to track movement objects in entire of frames on video.
However, there is no agreement to track multiple small nutriments in the video which has interference of hand-held camera and ocean waves.
In this paper, tuna nutriment tracking using trajectory mapping in application to aquaculture fish tank has been presented and demonstrated to be promising for interference video containing multiple small nutriment datasets.
We have demonstrated tuna nutriment tracking using trajectory mapping and the method consistently performs well on the interference video with good precision trajectory result.
We expect our approach to open the door for future work and to go beyond for feature extraction of ripple activity and focus on integrating tracking approach and ripple activity to be a decision to control fish feeding machine.

\end{document}